\documentclass[11pt]{article}
\usepackage[utf8]{inputenc}
\usepackage[T1]{fontenc}
\usepackage{lmodern}
\usepackage[margin=1in]{geometry}
\usepackage{amsmath,amssymb}
\usepackage{graphicx}
\usepackage{booktabs}
\usepackage{longtable}
\usepackage{array}
\usepackage{calc}
\usepackage{parskip}
\usepackage{float}     
\usepackage[hidelinks]{hyperref}
\usepackage{url}

\setkeys{Gin}{width=\linewidth,keepaspectratio}

\title{Phylogenetic signal in marine mammal and bird vocalizations
captured by audio foundation models: the limited benefit of
domain-specific pretraining}
\author{Víctor Rincón Yepes\footnote{Independent Researcher;
  rinvictor@gmail.com}}
\date{}

\begin{document}
\maketitle

\begin{abstract}
\noindent Whether acoustic similarity among species reflects their
evolutionary history is a fundamental question in comparative
bioacoustics. Classical spectral features have shown modest phylogenetic
signal in birds and frogs, but no study has tested whether large
pretrained audio models, which encode richer acoustic representations,
recover stronger phylogenetic structure.

Here, we apply Mantel tests to compare acoustic distance matrices
derived from four audio representations with a phylogenetic distance
matrix across 32 marine mammal species (1,754 recordings; Watkins Marine
Mammal Sound Database). We compare hand-crafted MFCC features (105
dimensions) with three audio foundation models: the Audio Spectrogram
Transformer (AST; 768d), Contrastive Language-Audio Pretraining (CLAP;
512d), and a BEATs encoder fine-tuned on bioacoustics by the Earth
Species Project (BEATs-bio; 768d).

Within the 26-species cetacean clade, where divergence times are well
resolved (2--34 Myr), the foundation-model embeddings recover a strong
phylogenetic signal (CLAP \(r = 0.82\), BEATs-bio \(r = 0.82\), AST
\(r = 0.74\); all \(p < 0.001\)), among the highest
acoustic--phylogenetic correlations reported for any taxon. Across the
full 32-species matrix, the correlation decreases to \(r \approx 0.40\)
(CLAP \(r = 0.410\), \(p < 0.001\)), largely because the constant
divergence time assigned to the deep cetacean--pinniped split dilutes
the signal. In contrast, hand-crafted MFCC features show no significant
correlation (\(r = 0.040\), \(p = 0.338\)).

The three foundation models did not differ significantly from one
another, as paired bootstrap confidence intervals for \(\Delta r\) all
overlapped zero. However, all three consistently outperformed MFCC. This
advantage remained after projecting every embedding to the same
dimensionality (PCA to the 105-dimensional MFCC space), indicating that
the improvement is not simply a consequence of the higher dimensionality
of the learned representations.

The observed signal is also not explained by the cetacean--pinniped
dichotomy, which accounts for only 4--6\% of the acoustic variance.
Likewise, it persists after a partial Mantel test controlling for
dominant frequency (\(r_{\mathrm{partial}} = 0.404\), \(p < 0.001\)),
retaining 97\% of the original \(R^2\). These results indicate that
foundation models capture structural properties of vocalizations beyond
simple pitch.

Extreme examples of acoustic convergence (Bearded Seal \(\times\)
Northern Right Whale; 90 Myr divergence, cosine distance 0.098) and
divergence (Leopard Seal \(\times\) Weddell Seal; 10 Myr divergence,
cosine distance 0.773) illustrate how ecological pressures and
lineage-specific evolution can decouple acoustic similarity from
phylogenetic relatedness.

To assess whether these findings generalize beyond marine mammals, we
repeated the analysis on 20 bird species (607 recordings across seven
orders) and additionally evaluated BirdNET, a classifier trained on
approximately 6,000 bird species. Using the time-calibrated bird
phylogeny of Jetz et al.~(2012), the general-purpose foundation models
again recovered strong phylogenetic signal (AST \(r = 0.55\), CLAP
\(r = 0.52\); \(p < 0.001\)). Surprisingly, neither BirdNET nor the
bioacoustic BEATs-bio model outperformed the general-purpose models
(\(r \approx 0.32\)--0.36), and the differences among foundation models
were not statistically significant, although all substantially exceeded
the MFCC baseline.

Overall, our results show that pretrained audio representations encode
evolutionary information across two independent radiations. They also
suggest that domain-specific pretraining is not required to recover
phylogenetic signal and may even reduce it. General-purpose audio models
performed at least as well as models specialized for bioacoustics or
birds. Finally, we discuss several possible explanations, including
differences in training objectives, model architecture, and
training-data diversity, and propose a controlled fine-tuning experiment
to disentangle their respective contributions.

\end{abstract}

\section{Introduction}\label{introduction}

Animal vocalizations are shaped by three things at once: evolutionary
history, ecological pressure, and individual learning. Separating them
is a recurring problem in comparative bioacoustics (Odom et al., 2021).
One prediction of the evolutionary account is easy to state. If
evolutionary history leaves an imprint on how species sound, then
species that sound alike should tend to be more closely related. That
imprint has a name, \emph{phylogenetic signal} (Blomberg \& Garland,
2002), and a standard test: the Mantel correlation between an acoustic
distance matrix and a phylogenetic distance matrix. It has been detected
in frogs (Gingras et al., 2013; r = 0.17 across 90 species), in birds
(Arato \& Fitch, 2021; 137 species), in estrildid finches (Shimizu et
al., 2023), and in gobiid fishes (Vasconcelos et al., 2021; r = 0.47).
In every one of those studies the acoustic distances came from
hand-crafted spectral features: fundamental frequency, spectral entropy,
spectral centroid, Mel-frequency cepstral coefficients (MFCCs).

The set of available representations has grown a lot since then. Large
pretrained audio models such as CLAP (Wu et al., 2023), AST (Gong et
al., 2021), and BEATs (Chen et al., 2023) map sound into
high-dimensional embedding spaces where semantically or structurally
similar recordings sit close together. On bioacoustic benchmarks they
clearly outperform MFCCs (Hagiwara et al., 2022; Miron et al., 2025).
What is unknown is whether the geometry these models impose on audio
also reflects evolutionary relationships. That is what we test.

Marine mammals are a convenient starting point. Divergence times across
Cetacea and Pinnipedia are well estimated, from about 2 to 90 million
years of separation (McGowen et al., 2020; Berta et al., 2018), and
their calls span an unusually wide acoustic range: infrasonic fin whale
pulses at 20 Hz, high-frequency dolphin clicks, and the elaborate
underwater songs of Weddell seals. The Watkins Marine Mammal Sound
Database (Watkins et al., 2004) provides a curated, freely available
collection across 32 species.

We use the Watkins data to ask whether acoustic distances from four
representations (an MFCC baseline; the general-purpose AST and CLAP; and
the bioacoustic BEATs-bio) correlate with phylogenetic distance. We
check whether any observed correlation just reflects dominant frequency,
using a partial Mantel test, and we inspect the species pairs that sit
furthest from the trend.

We then repeat the whole procedure on birds. Birds are a separate
radiation with a very different vocal biology, including widespread
vocal learning in oscine passerines (Arato \& Fitch, 2021). They also
let us ask a sharper question. Alongside the general-purpose and
bioacoustic models we add BirdNET (Kahl et al., 2021), a network trained
to classify around 6,000 bird species. If domain-specific pretraining is
what drives phylogenetic-signal recovery, a dedicated bird model should
be the best model on birds. It is not. Explaining that negative result
is the part of the paper that we found most worth thinking about.

\begin{center}\rule{0.5\linewidth}{0.5pt}\end{center}

\section{Methods}\label{methods}

\subsection{Acoustic data (marine
mammals)}\label{acoustic-data-marine-mammals}

We used the Watkins Marine Mammal Sound Database via HuggingFace
(\texttt{confit/wmms-parquet}), supplemented from the full Archive.org
release (\path{watkins-marine-mammal-sound-database-full-cuts}) for
species with fewer than 20 clips in the curated set. The final dataset
held 1,754 recordings across 32 species (Table S1), with at least 23
clips per species (median 55, maximum 114). Original sample rates were
preserved (1,000 to 40,000 Hz, most at 22,050 Hz) for dominant-frequency
estimation. Audio was resampled at each model's input stage.

\subsection{Acoustic data (birds)}\label{acoustic-data-birds}

For the cross-taxon replication we selected 20 bird species from seven
orders: Passeriformes, Cuculiformes, Bucerotiformes, Coraciiformes,
Accipitriformes, Columbiformes, and Charadriiformes (Table S2). The set
spans a wide range of divergence depths, from three congeneric
\emph{Turdus} thrushes to inter-order splits close to 80 Myr, without
leaving the well-characterised parts of the phylogeny. Focal recordings
came from the BirdCLEF 2023 metadata (Kahl et al., 2023) and were
downloaded from xeno-canto (www.xeno-canto.org), keeping only recordings
rated 2.5 or better. This left 607 recordings (15 to 35 per species,
median 35), resampled to 32 kHz.

\subsection{Audio representations}\label{audio-representations}

For every representation we computed one L2-normalised acoustic centroid
per species from the L2-normalised per-clip embeddings. Four
representations were applied to both taxa; BirdNET was applied only to
birds.

\textbf{MFCC baseline.} Each clip was summarised by 40 MFCCs (mean and
standard deviation over time) together with chroma, spectral contrast,
spectral centroid, spectral roll-off, and zero-crossing rate. This gave
a 105-dimensional feature vector, computed with \emph{librosa} 0.10
(McFee et al., 2015).

\textbf{AST.} We used \path{MIT/ast-finetuned-audioset-10-10-0.4593}
(Gong et al., 2021), a transformer trained on AudioSet (Gemmeke et al.,
2017) by supervised classification of audio-event tags. Audio was
resampled to 16 kHz and the pooler output gave a 768-dimensional
embedding per clip.

\textbf{CLAP.} We used \path{laion/clap-htsat-unfused} (Wu et al.,
2023), trained on 630,000 audio--text pairs by contrastive learning.
Audio was resampled to 48 kHz and the audio projection gave a
512-dimensional embedding.

\textbf{BEATs-bio.} We used
\path{EarthSpeciesProject/esp-aves2-naturelm-audio-v1-beats} (Miron et
al., 2025), the BEATs encoder (Chen et al., 2023) extracted from
NatureLM-audio after self-supervised fine-tuning on a large bioacoustic
corpus. Audio was resampled to 16 kHz and the mean over output frames
gave a 768-dimensional embedding.

\textbf{BirdNET (birds only).} We used BirdNET GLOBAL 6K v2.4 (Kahl et
al., 2021), a convolutional network trained to classify around 6,000
bird species. Audio was resampled to 48 kHz and processed in 3-second
windows. We read the 1,024-dimensional global average-pooling layer (the
layer immediately before the classifier) and averaged it over windows
for each clip. BirdNET is the only representation in our set trained by
supervised classification of the target group; the others are trained by
objectives that never see species labels.

Pairwise acoustic distances were cosine distances between L2-normalised
centroids, \(d_{ij} = 1 - \hat{c}_i \cdot \hat{c}_j\).

\subsection{Phylogenetic distances}\label{phylogenetic-distances}

For marine mammals we built the 32 \(\times\) 32 distance matrix from
published divergence times: McGowen et al.~(2020) for cetacean pairs,
Berta et al.~(2018) for pinniped pairs, and a fixed 90 Myr for
cetacean--pinniped pairs (the approximate Cetacea--Carnivora split; Dos
Reis et al., 2012). Distances range from 2 Myr for congenerics to 90
Myr, over 496 unique pairs.

For birds we used the time-calibrated phylogeny of Jetz et al.~(2012).
We pulled 100 trees from the posterior (Hackett backbone, ``all
species'' distribution) pruned to our 20 species through birdtree.org,
computed each tree's patristic distance matrix, converted it to time
since the most recent common ancestor (patristic distance divided by
two, since the trees are ultrametric), and averaged across the 100
trees. Three names were updated to their Jetz (2012) tip labels:
\emph{Chalcomitra amethystina} \(\rightarrow\) \emph{Nectarinia
amethystina}, \emph{Lophoceros nasutus} \(\rightarrow\) \emph{Tockus
nasutus}, and \emph{Turdus abyssinicus} \(\rightarrow\) \emph{Turdus
olivaceus}. Divergence times ran from 9 Myr for the congeneric thrushes
to 83 Myr, with a median of 81 Myr, so the sample is biased toward deep
splits (Table S2). As a sanity check, this molecular matrix correlates
at r = 0.96 with a simpler matrix calibrated by taxonomic rank
(following Prum et al., 2015); all results below use the Jetz matrix.

\subsection{Mantel tests}\label{mantel-tests}

For each representation we correlated the acoustic and phylogenetic
distance matrices with a Mantel test (Mantel, 1967): the Pearson
correlation between the upper triangles, with significance from 9,999
permutations of the row--column ordering. We computed 95\% confidence
intervals on each r by resampling species with replacement (2,000
iterations), and we compared models with a paired bootstrap on
\(\Delta r\) (resampling species jointly across the two matrices).
Because the representations span very different dimensionalities (105 to
1,024), and cosine distances behave differently in spaces of different
size, we repeated the entire analysis after projecting every embedding
to the MFCC dimensionality with PCA fit at the clip level.

\subsection{Partial Mantel test}\label{partial-mantel-test}

To ask whether the acoustic--phylogenetic correlation is really just
dominant frequency in disguise, we ran a partial Mantel test (Smouse et
al., 1986) controlling for a matrix of dominant-frequency distances.
Dominant frequency was the peak of the power spectrum of each clip
(ignoring energy below 30 Hz), summarised per species as the median on a
\(\log_2\) scale. Frequency distances were \(|\log_2 f_i - \log_2 f_j|\)
in octaves.

\begin{center}\rule{0.5\linewidth}{0.5pt}\end{center}

\section{Results}\label{results}

\subsection{The shape of the acoustic
space}\label{the-shape-of-the-acoustic-space}

Hierarchical clustering of the CLAP centroids produces a dendrogram
whose first split does not follow the textbook Mysticeti/Odontoceti
division (Figure 1a). Instead, species separate by frequency regime.
Low-frequency baleen whales cluster together; high-frequency dolphins
cluster together; and two pinnipeds, the bearded seal and the harp seal,
sit with the baleen whales rather than with the other seals. The UMAP
projection (Figure 1b; McInnes et al., 2018) shows the same thing more
geographically: a gradient ordered by call frequency, from the fin whale
at one extreme to the high-frequency dolphins at the other, with
pinnipeds scattered along it according to what they sound like rather
than what they are.

\begin{figure}[H]
\centering
\includegraphics[width=0.8\textwidth,height=\textheight]{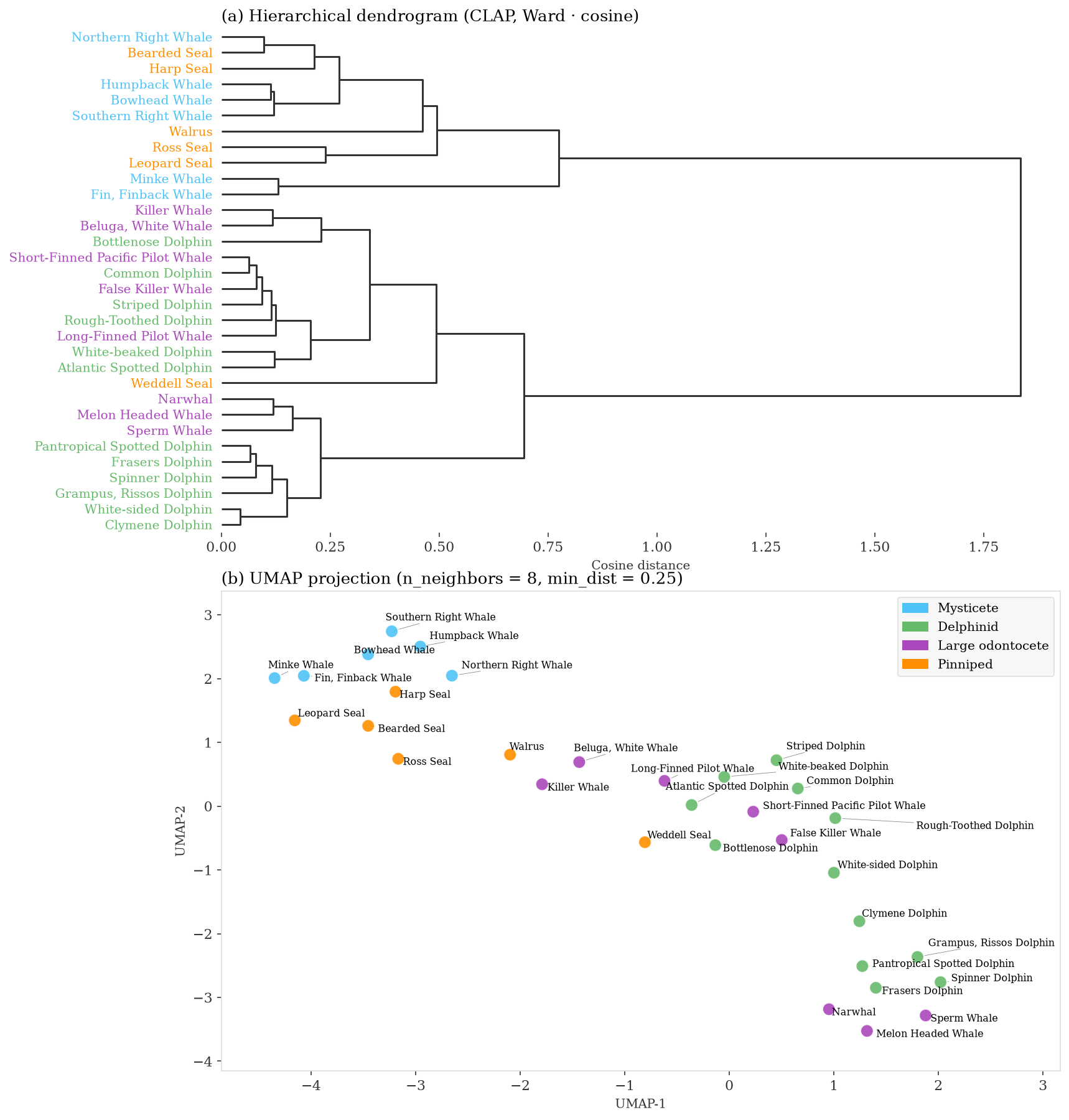}
\caption{Acoustic space of 32 marine mammal species in CLAP embeddings.
(a) Hierarchical dendrogram (Ward linkage, cosine distance on
L2-normalised centroids), coloured by taxonomic group. The first split
follows frequency, not the Mysticeti/Odontoceti division. (b) UMAP
projection of the same centroids (n\_neighbors = 8, min\_dist = 0.25,
cosine metric).}
\end{figure}

\subsection{Foundation models recover the signal; MFCC does
not}\label{foundation-models-recover-the-signal-mfcc-does-not}

Mantel results are summarised in Table 1 and shown in Figures 2 and 3.
MFCC features find no signal on the full 32-species matrix (r = 0.040, p
= 0.338). The three foundation models do: BEATs-bio r = 0.443, CLAP r =
0.410, AST r = 0.385 (all p \(\leq\) 0.003). The scatter plots (Figure
3) make the difference visible. For MFCC the cloud is flat. For the
foundation models cosine distance rises with divergence time, with the
largest step between within-clade pairs and the cetacean--pinniped pairs
at 90 Myr.

\begin{figure}[H]
\centering
\includegraphics[width=0.72\textwidth,height=\textheight]{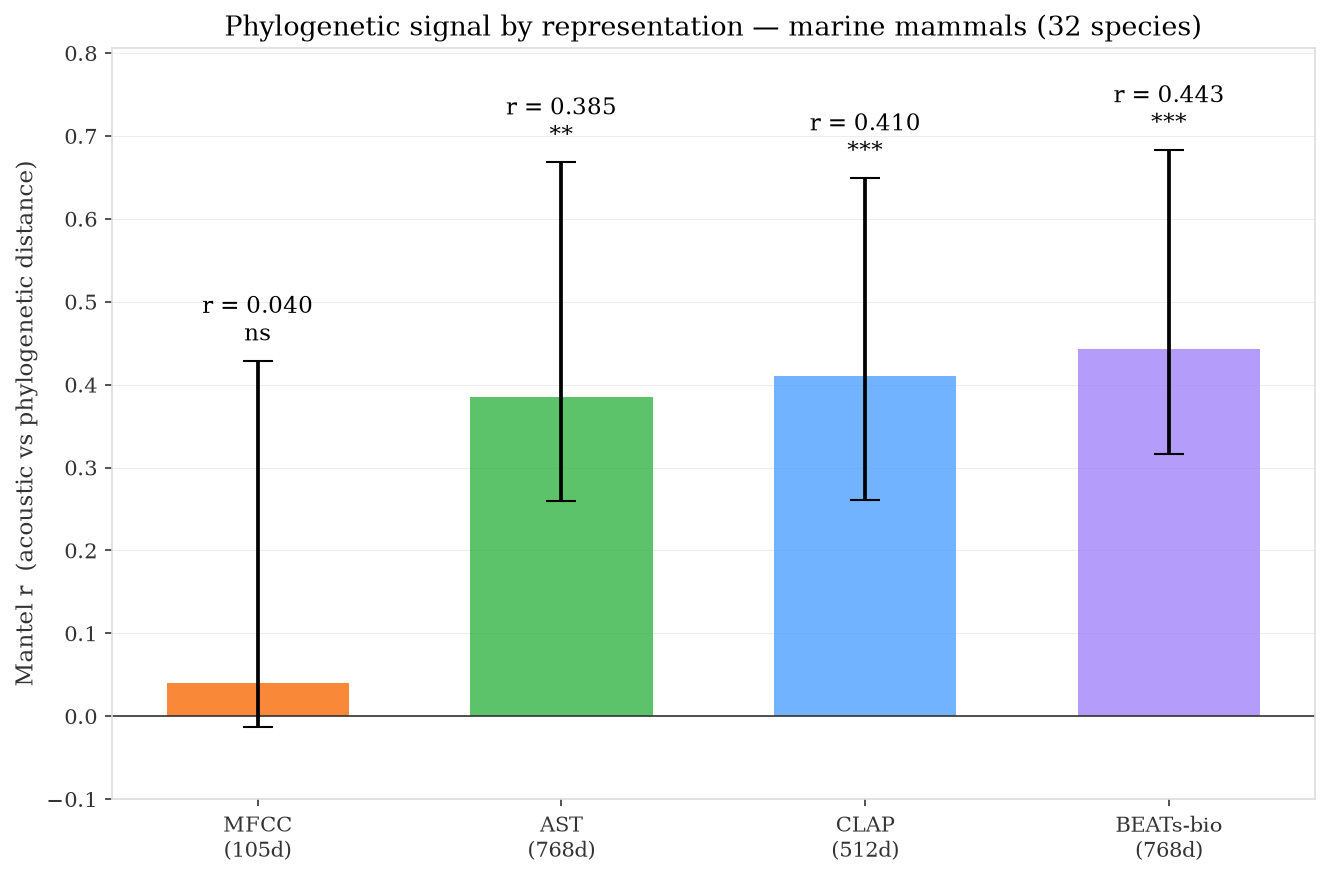}
\caption{Mantel r for each representation against phylogenetic distance
across 32 marine mammal species, with 95\% bootstrap confidence
intervals. Significance from 9,999 permutations (*** p \textless{}
0.001, ** p \textless{} 0.01, ns not significant). The three foundation
models are statistically indistinguishable from one another, but each
clearly exceeds MFCC.}
\end{figure}

\begin{table}[H]
\begingroup\centering
\begin{tabular}{@{}llllll@{}}
\toprule
Representation & Dimensions & r & p & R\(^2\) (\%) & 95\% CI \\
\midrule
MFCC & 105 & 0.040 & 0.338 & 0.2 & {[}-0.01, 0.43{]} \\
AST & 768 & 0.385 & 0.003 & 14.8 & {[}0.26, 0.67{]} \\
CLAP & 512 & 0.410 & \textless0.001 & 16.8 & {[}0.26, 0.65{]} \\
BEATs-bio & 768 & 0.443 & \textless0.001 & 19.6 & {[}0.32, 0.68{]} \\
\bottomrule
\end{tabular}
\par\endgroup
\smallskip
\textbf{Table 1.} Mantel results for four representations against phylogenetic distance across 32 marine mammal species (496 pairs). \emph{p-values from 9,999 permutations; 95\% confidence intervals from 2,000 species resamplings.}
\end{table}

The three foundation models are not distinguishable from one another at
this sample size. Their bootstrap intervals overlap heavily, and the
paired-bootstrap intervals on \(\Delta r\) all cross zero. What is
unambiguous is the gap between any foundation model and MFCC. That gap
also holds up under a matched-dimensionality control. After
PCA-projecting each foundation embedding down to 105 dimensions, the
Mantel r drops to between 0.20 and 0.26, while MFCC stays at 0.04. That
is still a five- to sixfold difference. So the foundation advantage is
not an artefact of embedding size. The absolute value of the correlation
does depend on dimensionality (it roughly halves under projection),
which implies that the signal is distributed across a subspace wider
than 105 dimensions.

\begin{figure}[H]
\centering
\includegraphics[width=0.8\textwidth,height=\textheight]{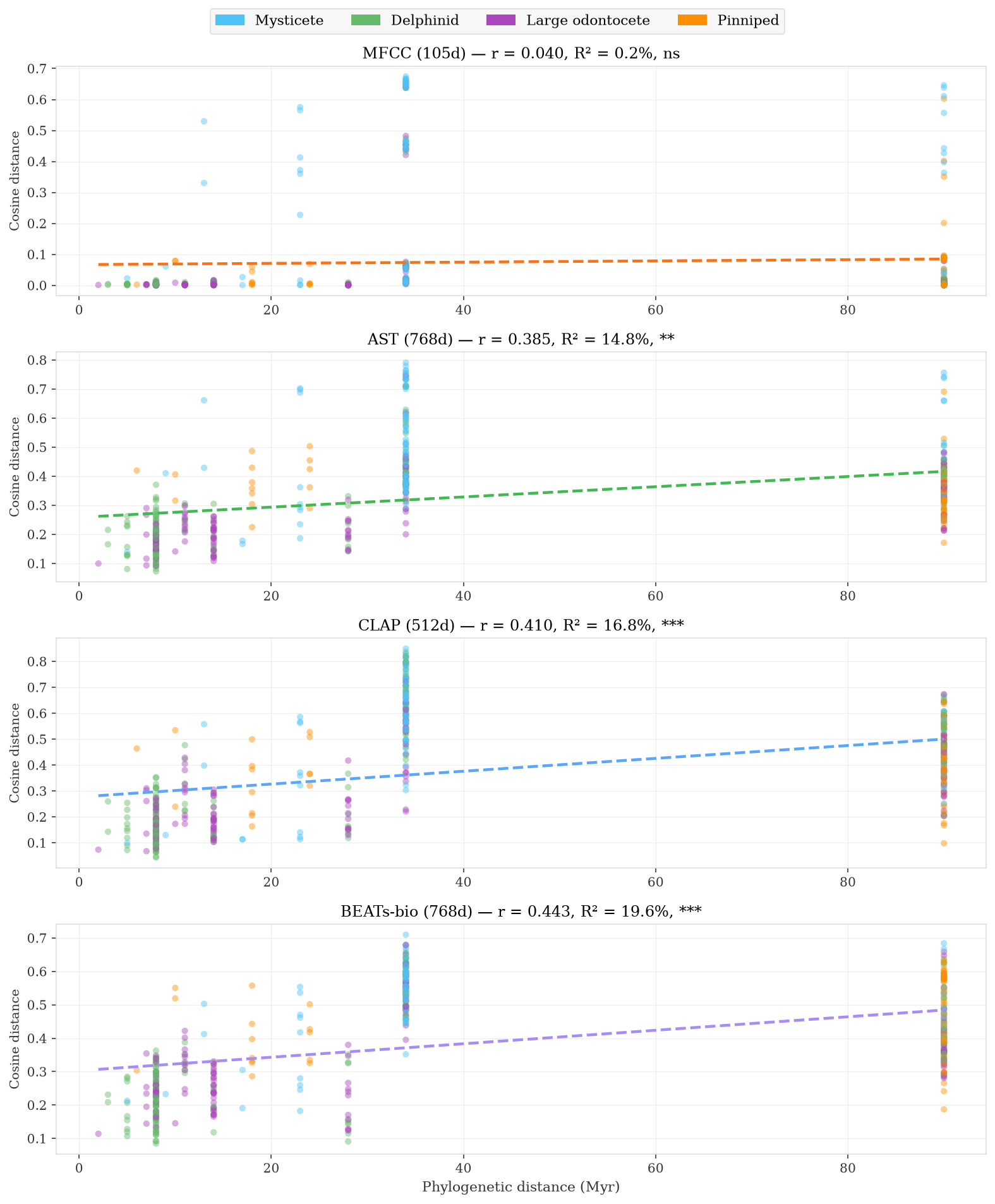}
\caption{Pairwise cosine distance against phylogenetic distance for each
representation (points coloured by taxonomic group; dashed line is the
least-squares fit). The MFCC cloud is flat; the three foundation models
show a positive slope.}
\end{figure}

\subsection{The signal is stronger within cetaceans, and it is not the
pinniped
split}\label{the-signal-is-stronger-within-cetaceans-and-it-is-not-the-pinniped-split}

Every cetacean--pinniped pair was assigned the same 90 Myr, so a large
block of the full matrix is constant. That raises an obvious concern:
perhaps the correlation only reflects ``cetaceans sound different from
pinnipeds,'' and nothing finer than that. Two checks argue against it.
First, a binary matrix encoding only the cetacean--pinniped split
accounts for 4 to 6\% of the acoustic variance across models, even
though it correlates almost perfectly with the phylogenetic matrix (r =
0.96). Second, restricting the analysis to the 26 cetaceans, which
removes the flat block entirely, strengthens the correlation rather than
weakening it: CLAP r = 0.82, AST r = 0.74, BEATs-bio r = 0.82 (all p
\textless{} 0.001). So the signal comes from fine phylogenetic structure
within a single clade, and the full-matrix values near r = 0.4 are
conservative, watered down by the constant 90 Myr rows.

Within cetaceans even MFCC becomes significant (r = 0.45, p \textless{}
0.001). Its failure on the full matrix, then, was partly a failure to
distinguish pinnipeds from cetaceans. The foundation models still lead
by a wide margin within the cetacean clade.

\subsection{Frequency does not explain
it}\label{frequency-does-not-explain-it}

A natural alternative story is that related species share body size,
that body size determines pitch, and that the models are largely
encoding pitch. The partial Mantel test disagrees. The bivariate
correlations of dominant frequency with CLAP and with phylogeny are both
weak (r = -0.085 and r = -0.133 respectively). Controlling for frequency
moves the CLAP--phylogeny correlation only slightly, from 0.410 to
0.404, retaining 97\% of the R\(^2\) (Figure 4). Whatever the foundation
models are picking up, it is not the dominant frequency of the call.

\begin{figure}[H]
\centering
\includegraphics[width=1\textwidth,height=\textheight]{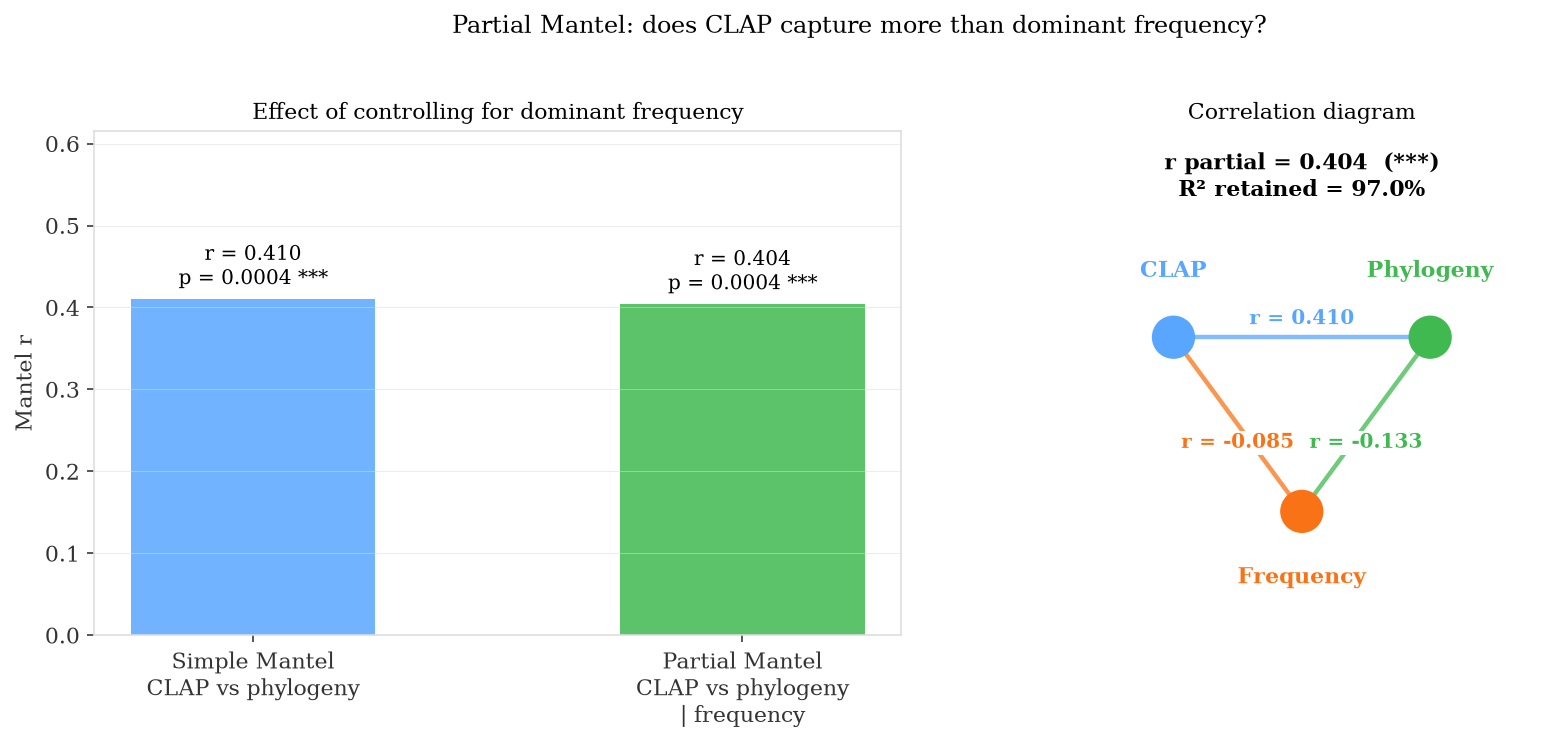}
\caption{Partial Mantel analysis controlling for dominant frequency. (a)
Simple Mantel r (CLAP vs phylogeny) and partial Mantel r (CLAP vs
phylogeny given dominant frequency): controlling for frequency retains
97\% of the R\(^2\) (r drops from 0.410 to 0.404). (b) The three
bivariate correlations; frequency is weakly related to both the acoustic
and the phylogenetic distances, so it is not a confound.}
\end{figure}

\subsection{Convergence and
divergence}\label{convergence-and-divergence}

The species pairs that fall furthest from the regression line are
biologically readable. The most convergent pair, closest in sound
despite being far apart in the tree, is the bearded seal and the
northern right whale: cosine distance 0.098 at 90 Myr apart. Bearded
seal calls are long, frequency-descending trills with rich harmonic
content (Risch et al., 2007), and in the embedding space they land right
next to baleen whale song. Three more cetacean--pinniped pairs appear
among the ten most convergent. All of them are low-frequency species,
and low-frequency Watkins recordings share a narrow recording bandwidth
(Section 4.3), so part of the closeness could reflect recording
conditions rather than the calls themselves. The biological reading is
supported by the acoustic literature we cite, but it should be treated
as a hypothesis to be checked with bandwidth-matched recordings.

The opposite pattern, close in the tree but far in sound, belongs to the
leopard seal and the Weddell seal: cosine distance 0.773 at only 10 Myr
apart. That is further apart in acoustic space than most
cetacean-to-cetacean pairs. Weddell seals produce elaborate underwater
calls year-round (Thomas \& Kuechle, 1982); leopard seals produce
sparse, low-frequency growls (Rogers, 2003). Two Antarctic phocids in
the same family have ended up at opposite ends of the acoustic space.

\subsection{The same pattern in birds, and a surprise about
domain-specific
models}\label{the-same-pattern-in-birds-and-a-surprise-about-domain-specific-models}

The bird analysis reproduces the main result and adds an unexpected
wrinkle (Figure 5, Table 2). Under the Jetz et al.~(2012) phylogeny the
general-purpose models again recover strong signal, with AST r = 0.549
and CLAP r = 0.521 (both p \textless{} 0.001, R\(^2\) near 30\%). Those
are actually higher than the marine mammal full-matrix values. MFCC is
again the weakest (r = 0.210, p = 0.029).

The unexpected part is what happens to the two domain-adapted models.
Neither BEATs-bio (r = 0.318) nor BirdNET (r = 0.361) beats the
general-purpose AST and CLAP. Both come in at roughly half the R\(^2\)
of the general-purpose ones. A paired bootstrap confirms that all four
foundation models beat MFCC (for example \(\Delta r\)(AST, MFCC) =
+0.36, 95\% CI {[}+0.20, +0.55{]}), but no two foundation models differ
significantly from each other, BirdNET included. A network trained
end-to-end to tell 6,000 bird species apart produced embeddings that
were no better, and by the point estimate slightly worse, at recovering
avian phylogeny than a transformer trained on general environmental
audio.

The signal survives the frequency control here as well (BirdNET r =
0.361 \(\rightarrow\) 0.343, retaining 90\% of the R\(^2\), p = 0.015).
Unlike in marine mammals, the within-clade signal is weak: restricted to
the 11 passerines the Mantel r ranges from -0.05 to 0.27 across models,
all with p \textgreater{} 0.09. This still holds under the molecular
phylogeny, so it is not an artefact of a coarse tree. Two things
probably compound. First, n = 11 is small. Second, oscine passerines
learn their songs, and vocal learning is expected to erode phylogenetic
signal at shallow depths (Arato \& Fitch, 2021). The avian signal, then,
rests mostly on the deep differences between orders.

\begin{figure}[H]
\centering
\includegraphics[width=0.8\textwidth,height=\textheight]{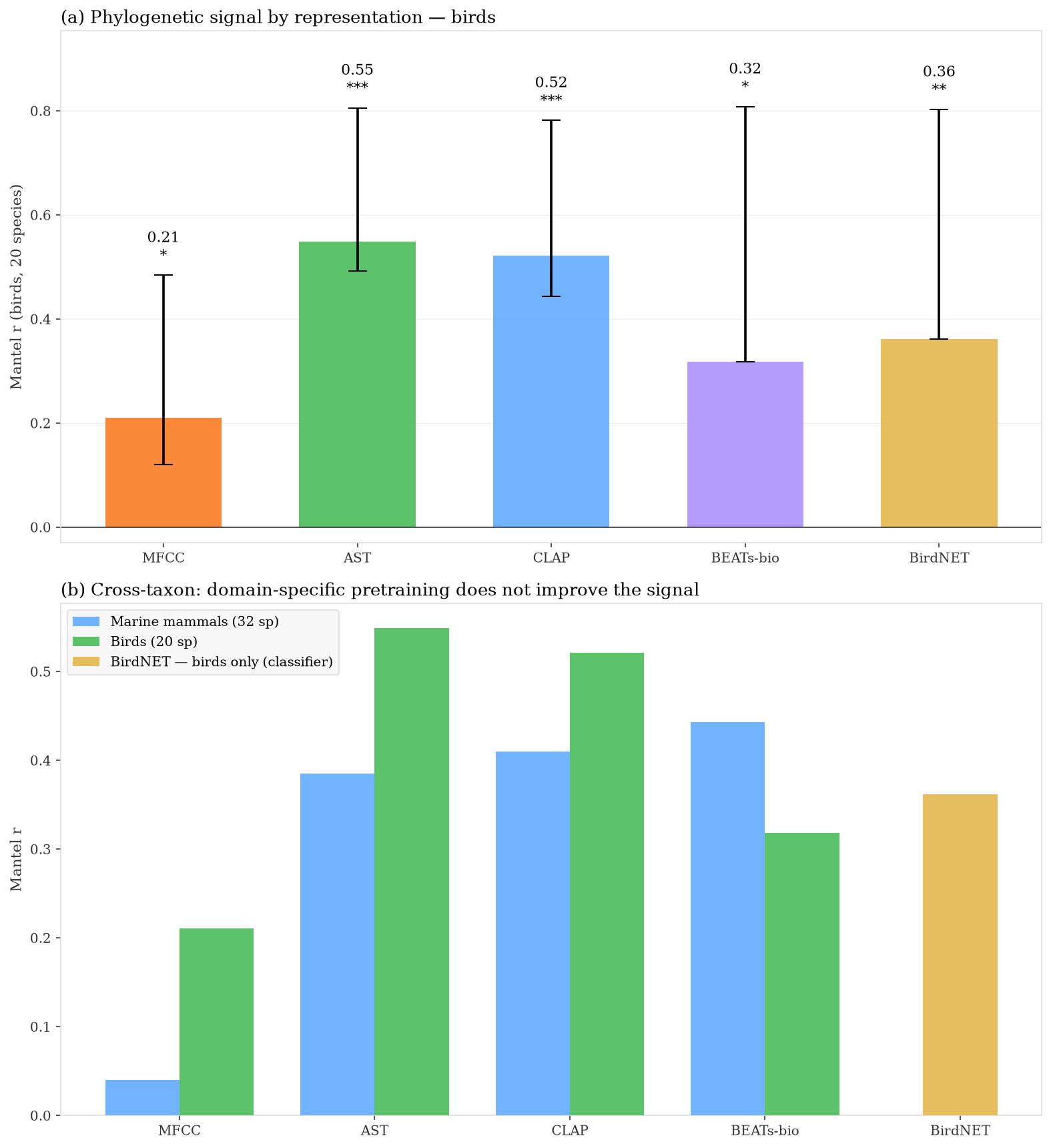}
\caption{Cross-taxon comparison. (a) Mantel r for five representations
against the Jetz et al.~(2012) phylogeny across 20 bird species, with
95\% bootstrap intervals (*** p \textless{} 0.001, ** p \textless{}
0.01, * p \textless{} 0.05). (b) Mantel r for marine mammals (32
species) and birds (20 species) side by side for the four shared
representations, plus the bird specialist BirdNET (birds only). The
general-purpose models lead in both groups; the domain-adapted models do
not.}
\end{figure}

\begin{table}[H]
\begingroup\centering
\begin{tabular}{@{}llllll@{}}
\toprule
Representation & Dimensions & Training objective & r & p & R\(^2\) (\%) \\
\midrule
MFCC & 102 & --- & 0.210 & 0.029 & 4.4 \\
AST & 768 & supervised (AudioSet tags) & 0.549 & \textless0.001 &
30.1 \\
CLAP & 512 & contrastive (audio--text) & 0.521 & \textless0.001 &
27.2 \\
BEATs-bio & 768 & self-supervised (bioacoustics) & 0.318 & 0.014 &
10.1 \\
BirdNET & 1024 & supervised classification (\textasciitilde6,000 birds)
& 0.361 & 0.009 & 13.1 \\
\bottomrule
\end{tabular}
\par\endgroup
\smallskip
\textbf{Table 2.} Mantel results for five representations against the Jetz et al.~(2012) phylogeny across 20 bird species (190 pairs). \emph{p-values from 9,999 permutations.}
\end{table}

\begin{center}\rule{0.5\linewidth}{0.5pt}\end{center}

\section{Discussion}\label{discussion}

\subsection{Learned representations carry evolutionary
information}\label{learned-representations-carry-evolutionary-information}

The headline result is that three off-the-shelf audio models, none of
them exposed to animal sound during training, recover significant
phylogenetic signal in marine mammal vocalizations, while hand-crafted
MFCC features do not. The effect is large within cetaceans (r near 0.8)
and reproduces in a second independent radiation, birds.

MFCCs missing the signal is not, in retrospect, a surprise. They
summarise local spectral texture at a fixed time scale and discard the
temporal shape of the call. The foundation models were trained to
discriminate sounds globally, and their embeddings appear to capture
something like the \emph{style} of a vocalization: call structure,
temporal modulation, and the shape of the spectral envelope. Roughly 15
to 30\% of the variance in acoustic distance travels with phylogenetic
distance under those representations. Inside the cetacean clade the
fraction is much higher.

\subsection{Domain-specific pretraining does not help, and we cannot say
why
yet}\label{domain-specific-pretraining-does-not-help-and-we-cannot-say-why-yet}

The bird experiment was designed to test the domain-specificity
hypothesis directly. The result was negative. BirdNET, trained
end-to-end to separate 6,000 bird species, did not beat the
general-purpose models on birds, and its point estimate came in lower.
BEATs-bio behaved similarly. Matching the training domain to the test
group is not by itself enough to improve phylogenetic-signal recovery.
In our setup it did not help.

Why the domain-adapted models score lower is not identifiable from four
models. Three explanations are compatible with what we see, and we
cannot rule any of them out.

The first is the training objective. BirdNET is optimised by supervised
species classification. That kind of pressure rewards collapsing exactly
the graded, within- and between-species similarity that a Mantel test
needs, pushing ``how similar do a thrush and a robin sound'' toward
``these are different classes.'' Calling this a supervision effect,
though, would be wrong. AST is also trained by supervised
classification, on AudioSet event tags, and it comes out the strongest
of all four models. If an objective effect exists, it is fine-grained
species discrimination specifically, not supervised training as a
category.

The second is architecture. BirdNET is a CNN. AST, CLAP, and BEATs are
transformers, whose attention may be better at capturing long-range
spectro-temporal structure than convolutions are. The awkward part is
that in our model set the only convolutional model is also the only
species classifier, so we cannot separate an architecture effect from an
objective effect. This is a real confound in the model list, not just in
the interpretation.

The third is training-data breadth. The two strongest models were
trained on broad general audio (AudioSet, audio--text pairs on the open
web). The two weaker models were trained on narrower, specialised
corpora. Squeezing the training distribution may cost some of the
representational richness that happens to correlate with phylogeny.

What we can defend is the negative claim: domain-specific pretraining
did not improve signal recovery in either taxon, and in birds it
slightly reduced it. Separating the three explanations needs an
experiment that holds architecture and data fixed and varies only the
objective. Take one pretrained encoder, AST for example, and compare the
phylogenetic signal of its embeddings before and after fine-tuning it
with a species-classification head. If the objective is the cause, the
signal should drop under fine-tuning. If it does not, architecture and
data become the more likely story. We did not run that experiment and we
flag it as the natural next step.

\subsection{Limitations}\label{limitations}

Several caveats apply. The marine mammal recordings come from a single
historical archive spanning several decades and a range of equipment.
Recording conditions are partly bound up with taxonomy, since clips of
related species tend to come from the same expeditions, so sample rate
and noise floor travel with phylogeny. The modal recording sample rate,
expressed as an octave distance, correlates with phylogenetic distance
at r = 0.36 and with the embedding distances at 2 to 5\% of variance.
Our partial Mantel controlled for dominant frequency but not for sample
rate. That said, the within-cetacean signal (r near 0.8) is much larger
than any plausible recording effect could produce. All models tested
were trained on airborne audio, not underwater sound, and a model
trained on hydrophone data might do better still.

The bird analysis is a cross-taxon replication, not a full avian study.
We used 20 species biased toward deep splits, so the within-order signal
is underpowered and the ordering among foundation models is not
resolved. The two claims we stand behind are narrow. Every foundation
model beats MFCC. The domain-specialist BirdNET does not beat the
general-purpose models. Nailing down the mechanism will need more
species, denser sampling within orders, and the fine-tuning experiment
described in Section 4.2.

\subsection{Practical implications}\label{practical-implications}

For anyone selecting an embedding for comparative work, the practical
takeaway is not what we expected going in. Any of the three
general-purpose foundation models will beat hand-crafted MFCCs by a wide
margin. A model advertised for your taxon, on the other hand, is not
automatically the better choice for phylogenetically structured
analysis. In our setup, in two taxa, it was not.

\begin{center}\rule{0.5\linewidth}{0.5pt}\end{center}

\section{Conclusions}\label{conclusions}

Audio foundation models recover phylogenetic signal from vocalizations
in two independent radiations, marine mammals and birds. Classical MFCC
features do so only weakly, and only within a clade. In marine mammals
the signal is very strong within cetaceans, survives control for the
pinniped split, survives control for dominant frequency, and does not
depend on embedding dimensionality. In birds it is strong at the level
of orders, consistent under the Jetz et al.~(2012) phylogeny, and again
largely independent of the model's training domain.

The cross-taxon comparison delivers the least expected finding.
Domain-specific pretraining is neither necessary nor sufficient for
strong signal recovery. A bird classifier and a bioacoustic model did no
better than general-purpose transformers, and by the point estimate
slightly worse. Which factor is responsible (training objective,
architecture, or the breadth of the training data) cannot be identified
from currently available models, and we sketch the controlled experiment
that would tell them apart. For now, the safe reading is that
general-purpose audio embeddings are a practical tool for comparative
bioacoustics, and that the domain label on a model is not, by itself, a
reason to prefer it for this task.

\begin{center}\rule{0.5\linewidth}{0.5pt}\end{center}

\section{Data and code availability}\label{data-and-code-availability}

Marine mammal audio is public from the Watkins database (HuggingFace
\texttt{confit/wmms-parquet}; Archive.org
\path{watkins-marine-mammal-sound-database-full-cuts}); bird audio is
from xeno-canto, selected through the BirdCLEF 2023 metadata. Processed
embeddings, distance matrices, and all analysis code are available at
\url{https://github.com/rinvictor/bioacoustic-phylogeny-embeddings}.
Models: CLAP \path{laion/clap-htsat-unfused}; AST
\path{MIT/ast-finetuned-audioset-10-10-0.4593}; BEATs-bio
\path{EarthSpeciesProject/esp-aves2-naturelm-audio-v1-beats}; BirdNET
GLOBAL 6K v2.4 (via \texttt{birdnetlib}). Supplementary Tables S1
(marine mammals) and S2 (birds) list the species, counts, and sources.

\begin{center}\rule{0.5\linewidth}{0.5pt}\end{center}

\section{References}\label{references}

Arato, J., \& Fitch, W. T. (2021). Phylogenetic signal in the
vocalizations of vocal learning and vocal non-learning birds.
\emph{Philosophical Transactions of the Royal Society B}, 376, 20200241.

Berta, A., Churchill, M., \& Boessenecker, R. W. (2018). The origin and
evolutionary biology of pinnipeds: seals, sea lions, and walruses.
\emph{Annual Review of Earth and Planetary Sciences}, 46, 203--228.

Blomberg, S. P., \& Garland, T. (2002). Tempo and mode in evolution:
phylogenetic inertia, adaptation and comparative methods. \emph{Journal
of Evolutionary Biology}, 15, 899--910.

Chen, S., Wu, Y., Wang, C., Liu, S., Tompkins, D., Chen, Z., \& Wei, F.
(2023). BEATs: Audio pre-training with acoustic tokenizers. In
\emph{Proceedings of ICML 2023}.

Dos Reis, M., Inoue, J., Hasegawa, M., Asher, R. J., Donoghue, P. C., \&
Yang, Z. (2012). Phylogenomic datasets provide both precision and
accuracy in estimating the timescale of placental mammal phylogeny.
\emph{Proceedings of the Royal Society B}, 279, 3491--3500.

Gemmeke, J. F., Ellis, D. P. W., Freedman, D., Jansen, A., Lawrence, W.,
Moore, R. C., \ldots{} Ritter, M. (2017). Audio Set: An ontology and
human-labeled dataset for audio events. In \emph{Proceedings of ICASSP
2017}.

Gingras, B., Boeckle, M., Herbst, C. T., \& Fitch, W. T. (2013). Call
acoustics reflect body size across four clades of anurans. \emph{BMC
Evolutionary Biology}, 13, 143.

Gong, Y., Chung, Y.-A., \& Glass, J. (2021). AST: Audio Spectrogram
Transformer. In \emph{Proceedings of Interspeech 2021}.

Hagiwara, M., Robinson, D., Meire, E., Newby, J., Keranen, K., \&
Spracklen, J. (2022). BEANS: The Benchmark of Animal Sounds.
\emph{arXiv:2210.12300}.

Jetz, W., Thomas, G. H., Joy, J. B., Hartmann, K., \& Mooers, A. O.
(2012). The global diversity of birds in space and time. \emph{Nature},
491, 444--448.

Kahl, S., Wood, C. M., Eibl, M., \& Klinck, H. (2021). BirdNET: A deep
learning solution for avian diversity monitoring. \emph{Ecological
Informatics}, 61, 101236.

Kahl, S., Denton, T., Klinck, H., Reers, H., Cherutich, F., Glotin, H.,
\ldots{} Vellinga, W.-P. (2023). Overview of BirdCLEF 2023: Automated
bird species identification in Eastern Africa. \emph{Working Notes of
CLEF 2023}.

Mantel, N. (1967). The detection of disease clustering and a generalized
regression approach. \emph{Cancer Research}, 27, 209--220.

McFee, B., Raffel, C., Liang, D., Ellis, D. P. W., McVicar, M.,
Battenberg, E., \& Nieto, O. (2015). librosa: Audio and music signal
analysis in Python. In \emph{Proceedings of the 14th Python in Science
Conference}.

McGowen, M. R., Tsagkogeorga, G., Álvarez-Carretero, S., Dos Reis, M.,
Struebig, M., Deaville, R., \ldots{} Rossiter, S. J. (2020).
Phylogenomic resolution of the cetacean tree of life using target
sequence capture. \emph{Systematic Biology}, 69, 479--501.

McInnes, L., Healy, J., \& Melville, J. (2018). UMAP: Uniform manifold
approximation and projection for dimension reduction.
\emph{arXiv:1802.03426}.

Miron, M., Robinson, D., Alizadeh, M., Gilsenan-McMahon, E., Narula, G.,
Chemla, E., \ldots{} Geist, M. (2025). What matters for bioacoustic
encoding. In \emph{Proceedings of ICLR 2025}.

Odom, K. J., Araya-Salas, M., Morano, J. L., Ligon, R. A., Leighton, G.
M., Taff, C. C., \ldots{} Wright, T. F. (2021). Comparative
bioacoustics: A roadmap for quantifying and comparing animal sounds
across diverse taxa. \emph{Biological Reviews}, 96, 1958--1989.

Prum, R. O., Berv, J. S., Dornburg, A., Field, D. J., Townsend, J. P.,
Lemmon, E. M., \& Lemmon, A. R. (2015). A comprehensive phylogeny of
birds (Aves) using targeted next-generation DNA sequencing.
\emph{Nature}, 526, 569--573.

Risch, D., Clark, C. W., Corkeron, P. J., Elepfandt, A., Kovacs, K. M.,
Lydersen, C., Stirling, I., \& Van Parijs, S. M. (2007). Vocalizations
of male bearded seals, \emph{Erignathus barbatus}: classification and
geographical variation. \emph{Animal Behaviour}, 73, 747--762.

Rogers, T. L. (2003). Factors influencing the acoustic behaviour of male
phocid seals. \emph{Aquatic Mammals}, 29, 247--260.

Shimizu, T., Ota, K., Suzuki, R., \& Arita, T. (2023). Acoustic
structure of vocalizations reflects phylogenetic relationships in
estrildid finches. \emph{Scientific Reports}, 13, 6832.

Smouse, P. E., Long, J. C., \& Sokal, R. R. (1986). Multiple regression
and correlation extensions of the Mantel test of matrix correspondence.
\emph{Systematic Zoology}, 35, 627--632.

Thomas, J. A., \& Kuechle, V. B. (1982). Quantitative analysis of
Weddell seal underwater vocalizations at McMurdo Sound, Antarctica.
\emph{Journal of the Acoustical Society of America}, 72, 1730--1738.

Vasconcelos, R. P., Pratas, J., Batista, V., Costa, M. J., \& Cabral, H.
N. (2021). Acoustic and genetic distances in gobiid fishes. \emph{PLOS
ONE}, 16, e0260810.

Watkins, W. A., Daher, M. A., Fristrup, K. M., Howald, T. J., \& di
Sciara, G. N. (2004). Marine mammal sound database. \emph{Marine Mammal
Science}, 20(1), 13--26.

Wu, Y., Chen, K., Zhang, T., Hui, Y., Berg-Kirkpatrick, T., \& Dubnov,
S. (2023). Large-scale contrastive language-audio pretraining with
feature fusion and keyword-to-caption augmentation. In \emph{Proceedings
of ICASSP 2023}.

\begin{center}\rule{0.5\linewidth}{0.5pt}\end{center}

\section{Supplementary Tables}\label{supplementary-tables}

Species are ordered by taxonomic group and common name. HF = Watkins
Marine Mammal Sound Database curated subset (HuggingFace,
\texttt{confit/wmms-parquet}); FA = the same database's full archive
(Archive.org, \path{watkins-marine-mammal-sound-database-full-cuts}),
used to supplement species with fewer than 20 clips in the curated set.
Dominant frequency is the median peak of the power spectrum across clips
(Hz); values marked † are likely underestimates because the historical
Watkins recordings for these species were sampled at 1,000--5,000 Hz
(Nyquist limit 500--2,500 Hz). Cetacean divergence times are from
McGowen et al.~(2020), pinniped divergence times from Berta et
al.~(2018), and the cetacean--pinniped separation is fixed at 90 Myr
(Dos Reis et al., 2012).

\emph{Mysticeti (baleen whales).}

{\small
\begin{longtable}[]{@{}lllll@{}}
\toprule\noalign{}
Common name & Scientific name & n clips & Dom. freq. (Hz) & Source \\
\midrule\noalign{}
\endhead
\bottomrule\noalign{}
\endlastfoot
Bowhead Whale & \emph{Balaena mysticetus} & 60 & 224 & HF \\
Fin Whale & \emph{Balaenoptera physalus} & 50 & 53 & HF \\
Humpback Whale & \emph{Megaptera novaeangliae} & 64 & 70 & HF \\
Minke Whale & \emph{Balaenoptera acutorostrata} & 23 & 75 & HF+FA \\
Northern Right Whale & \emph{Eubalaena glacialis} & 54 & 165 & HF \\
Southern Right Whale & \emph{Eubalaena australis} & 25 & 137 & HF \\
\end{longtable}
}

\emph{Delphinidae (oceanic dolphins).}

{\small
\begin{longtable}[]{@{}lllll@{}}
\toprule\noalign{}
Common name & Scientific name & n clips & Dom. freq. (Hz) & Source \\
\midrule\noalign{}
\endhead
\bottomrule\noalign{}
\endlastfoot
Atlantic Spotted Dolphin & \emph{Stenella frontalis} & 58 & 7,468 &
HF \\
Bottlenose Dolphin & \emph{Tursiops truncatus} & 24 & 6,490 & HF \\
Clymene Dolphin & \emph{Stenella clymene} & 63 & 208 & HF \\
Common Dolphin & \emph{Delphinus delphis} & 52 & 180 & HF \\
Fraser's Dolphin & \emph{Lagenodelphis hosei} & 87 & 683 & HF \\
Risso's Dolphin & \emph{Grampus griseus} & 67 & 849 & HF \\
Pantropical Spotted Dolphin & \emph{Stenella attenuata} & 66 & 2,588 &
HF \\
Rough-toothed Dolphin & \emph{Steno bredanensis} & 50 & 59 † & HF \\
Spinner Dolphin & \emph{Stenella longirostris} & 114 & 60 † & HF \\
Striped Dolphin & \emph{Stenella coeruleoalba} & 81 & 180 & HF \\
White-beaked Dolphin & \emph{Lagenorhynchus albirostris} & 57 & 60 † &
HF \\
White-sided Dolphin & \emph{Lagenorhynchus acutus} & 55 & 60 † & HF \\
\end{longtable}
}

\emph{Non-delphinid Odontoceti (other toothed whales).}

{\small
\begin{longtable}[]{@{}lllll@{}}
\toprule\noalign{}
Common name & Scientific name & n clips & Dom. freq. (Hz) & Source \\
\midrule\noalign{}
\endhead
\bottomrule\noalign{}
\endlastfoot
Beluga & \emph{Delphinapterus leucas} & 50 & 44 † & HF \\
False Killer Whale & \emph{Pseudorca crassidens} & 59 & 6,776 & HF \\
Killer Whale & \emph{Orcinus orca} & 35 & 781 & HF \\
Long-finned Pilot Whale & \emph{Globicephala melas} & 70 & 52 † & HF \\
Melon-headed Whale & \emph{Peponocephala electra} & 59 & 108 & HF \\
Narwhal & \emph{Monodon monoceros} & 50 & 60 † & HF \\
Short-finned Pilot Whale & \emph{Globicephala macrorhynchus} & 67 &
1,915 & HF \\
Sperm Whale & \emph{Physeter macrocephalus} & 75 & 174 & HF \\
\end{longtable}
}

\emph{Pinnipedia (seals and walrus).}

\begin{table}[H]
\begingroup\centering
{\small
\begin{tabular}{@{}lllll@{}}
\toprule
Common name & Scientific name & n clips & Dom. freq. (Hz) & Source \\
\midrule
Bearded Seal & \emph{Erignathus barbatus} & 37 & 319 & HF \\
Harp Seal & \emph{Pagophilus groenlandicus} & 47 & 42 † & HF \\
Leopard Seal & \emph{Hydrurga leptonyx} & 25 & 352 & HF+FA \\
Ross Seal & \emph{Ommatophoca rossii} & 50 & 325 & HF \\
Walrus & \emph{Odobenus rosmarus} & 38 & 341 & HF \\
Weddell Seal & \emph{Leptonychotes weddellii} & 42 & 120 & HF+FA \\
\bottomrule
\end{tabular}
}
\par\endgroup
\smallskip
\textbf{Table S1.} The 32 marine mammal species included in the analysis (32 species; 1,754 clips; n per species: min 23, median 55, max 114).
\end{table}

Focal bird recordings came from the BirdCLEF 2023 metadata (Kahl et al.,
2023) downloaded from xeno-canto (quality rating \(\geq\) 2.5). Pairwise
divergence times come from the time-calibrated phylogeny of Jetz et
al.~(2012): 100 posterior Hackett-backbone trees pruned via birdtree.org
and averaged as time since the most recent common ancestor. Divergence
times span 9 Myr (congeneric \emph{Turdus}) to 83 Myr (inter-order),
median 81 Myr; the sample is weighted toward deep divergences. † =
species mapped to a different Jetz (2012) tip label.

\begin{table}[H]
\begingroup\centering
{\small
\begin{tabular}{@{}lllll@{}}
\toprule
Common name & Scientific name & Family & Order & n \\
\midrule
Barn Swallow & \emph{Hirundo rustica} & Hirundinidae & Passeriformes &
35 \\
African Paradise Flycatcher & \emph{Terpsiphone viridis} & Monarchidae &
Passeriformes & 35 \\
African Pied Wagtail & \emph{Motacilla aguimp} & Motacillidae &
Passeriformes & 35 \\
African Dusky Flycatcher & \emph{Muscicapa adusta} & Muscicapidae &
Passeriformes & 22 \\
Amethyst Sunbird & \emph{Chalcomitra amethystina} † & Nectariniidae &
Passeriformes & 35 \\
African Black-headed Oriole & \emph{Oriolus larvatus} & Oriolidae &
Passeriformes & 35 \\
Baglafecht Weaver & \emph{Ploceus baglafecht} & Ploceidae &
Passeriformes & 23 \\
Blue-headed Crested Flycatcher & \emph{Elminia longicauda} &
Stenostiridae & Passeriformes & 18 \\
Abyssinian Thrush & \emph{Turdus abyssinicus} † & Turdidae &
Passeriformes & 26 \\
African Thrush & \emph{Turdus pelios} & Turdidae & Passeriformes & 35 \\
Bare-eyed Thrush & \emph{Turdus tephronotus} & Turdidae & Passeriformes
& 15 \\
African Emerald Cuckoo & \emph{Chrysococcyx cupreus} & Cuculidae &
Cuculiformes & 35 \\
Black-and-white-casqued Hornbill & \emph{Bycanistes subcylindricus} &
Bucerotidae & Bucerotiformes & 34 \\
African Grey Hornbill & \emph{Lophoceros nasutus} † & Bucerotidae &
Bucerotiformes & 35 \\
Blue-cheeked Bee-eater & \emph{Merops persicus} & Meropidae &
Coraciiformes & 35 \\
African Goshawk & \emph{Accipiter tachiro} & Accipitridae &
Accipitriformes & 35 \\
African Fish Eagle & \emph{Haliaeetus vocifer} & Accipitridae &
Accipitriformes & 35 \\
African Mourning Dove & \emph{Streptopelia decipiens} & Columbidae &
Columbiformes & 32 \\
African Green Pigeon & \emph{Treron calvus} & Columbidae & Columbiformes
& 23 \\
African Jacana & \emph{Actophilornis africanus} & Jacanidae &
Charadriiformes & 29 \\
\bottomrule
\end{tabular}
}
\par\endgroup
\smallskip
\textbf{Table S2.} The 20 bird species used for the avian replication (20 species; 607 recordings; 7 orders; n per species: min 15, median 35, max 35). Renamings for Jetz (2012) tip labels: \emph{Chalcomitra amethystina} \(\rightarrow\) \emph{Nectarinia amethystina}; \emph{Lophoceros nasutus} \(\rightarrow\) \emph{Tockus nasutus}; \emph{Turdus abyssinicus} \(\rightarrow\) \emph{Turdus olivaceus}.
\end{table}

\end{document}